\documentclass[10pt,twocolumn,letterpaper]{article}

\usepackage{cvm}
\usepackage{times}
\usepackage{epsfig}
\usepackage{graphicx}
\usepackage{bm}
\usepackage{subfig}
\usepackage{amsmath,amssymb,amsfonts}
\usepackage{algorithmic}
\usepackage{xcolor}
\usepackage{graphicx}
\usepackage{mathtools}
\usepackage{comment}
\usepackage{arydshln}
\usepackage[normalem]{ulem}
\useunder{\uline}{\ul}{}
\newcommand{\re}[1]{{\textcolor{black}{#1}}}

\usepackage[pagebackref=true,breaklinks=true,letterpaper=true,colorlinks,bookmarks=false]{hyperref}

\cvmfinalcopy

\def\cvmPaperID{****} 

\ifcvmfinal\fi
\begin{document}

\title{Learning to Predict Diverse Human Motions \\ from a Single Image via Mixture Density Networks}

\author{Chunzhi Gu\\
University of Fukui\\
Fukui, Japan\\
{\tt\small gchunzhi@u-fukui.ac.jp}
\and
Yan Zhao\\
Jiangsu University\\
Jiangsu, China\\
{\tt\small yanzhao\_cs@ujs.edu.cn}
\and
Chao Zhang\\
University of Fukui\\
Fukui, Japan\\
{\tt\small zhang@u-fukui.ac.jp}}

\maketitle

\begin{abstract}
Human motion prediction, which plays a key role in computer vision, generally requires a past motion sequence as input. However, in real applications, a complete and correct past motion sequence can be too expensive to achieve. In this paper, we propose a novel approach to predicting future human motions from a much weaker condition, i.e., a single image, with mixture density networks (MDN) modeling. Contrary to most existing deep human motion prediction approaches, the multimodal nature of MDN enables the generation of diverse future motion hypotheses, which well compensates for the strong stochastic ambiguity aggregated by the single input and human motion uncertainty. In designing the loss function, we further introduce the energy-based formulation to flexibly impose prior losses over the learnable parameters of MDN to maintain motion coherence as well as improve the prediction accuracy by customizing the energy functions. Our trained model directly takes an image as input and generates multiple plausible motions that satisfy the given condition. Extensive experiments on two standard benchmark datasets demonstrate the effectiveness of our method in terms of prediction diversity and accuracy.
\end{abstract}

\section{Introduction}
Human motion prediction has gained tremendous attention in recent years due to increasing needs in the field of human-related future forecasting tasks, such as human tracking \cite{gong2011multi} and autonomous driving \cite{paden2016survey}. Understanding human dynamics is challenging because of complicated biochemical body motions and a high degree of inherent uncertainties. As a result, the task of human motion prediction is generally cast as a sequence-to-sequence mapping problem that requires informative past observed human poses. By leveraging powerful temporal information encoders (e.g., hierarchical convolutional filters \cite{li2018convolutional}), a reliable future sequence can thus be recovered with Recurrent Neural Networks (RNN). 

\begin{figure}[t]
\begin{center}
\includegraphics[width=0.9\linewidth]{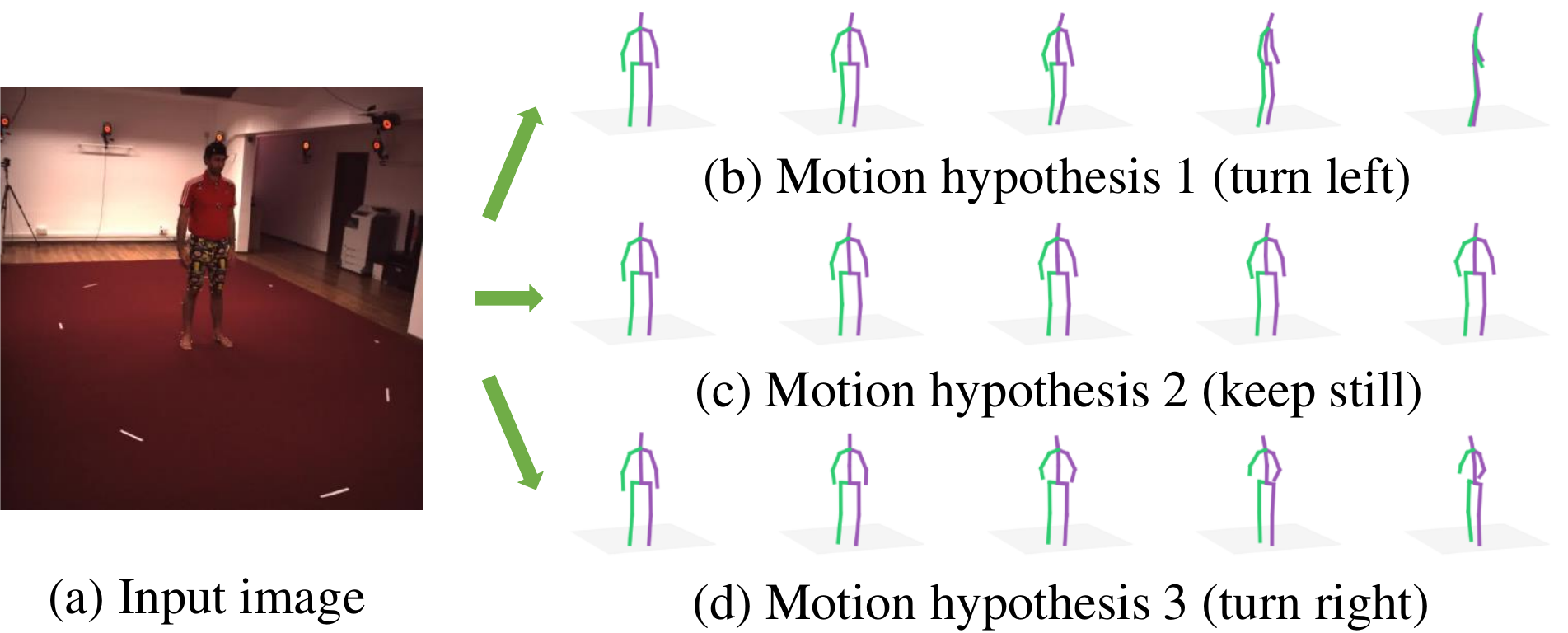}
\end{center}
\caption{Diverse future motion hypotheses ((b) $\sim$ (d)) generated from a single input image (a) via our MDN modeling. }
\label{fig:introduction}
\end{figure}

However, in real applications, it is usually the case that \re{the} easily accessible input is not informative enough to yield accurate future motion predictions, such as a single image. A pioneering work by \cite{chao2017forecasting} first attempts to address this problem by recurrently regressing \re{2D human skeletons with a sequence generator, followed by a 3D converter to generate 3D human poses.} Despite its novel concept, the whole system is deterministic, which means the network is not able to capture the potentially diverse future motions. Fig. \ref{fig:introduction} illustrates an example of multiple possible future motions induced by a single image. Without past motions as reference, a standing human (Fig. \ref{fig:introduction}(a)) in the image can be equally possible to turn left (Fig. \ref{fig:introduction}(b)), turn right (Fig. \ref{fig:introduction}(d)) or even keep still (Fig. \ref{fig:introduction}(c)) in the following frames. Consequently, various challenges imposed by the inherent pose and motion stochasticity suggest that a powerful model should be capable of modeling such inner ambiguities. We thus argue that contrary to a deterministic prediction system, it is more acceptable to produce multiple plausible future motions that satisfy the given image, as shown in Fig. \ref{fig:introduction}. This idea can manifest its essentiality in many safety-critical applications. For example, for autonomous vehicles, it is crucial to foresee the future of a straight walking pedestrian suddenly turning left, even though he will most probably remain walking in the same direction. 

In this paper, we resort to the mixture density networks (MDN) \cite{bishop1994mixture} to generate diverse human motion predictions. Our network architecture is a simple composition of an image to 2D estimation block, a 2D feature extractor block, and a 3D mixture density block, which straightforwardly takes a single image as input. Based on the features extracted by the deep neural networks, the mixture density block outputs the parameters for the mixture distribution, including several sets of means, standard deviations, and mixture coefficients. Specifically, we extend the mean of each distribution to represent a motion sequence in 3D space. Moreover, we design energy functions such that the involved prior loss can be flexibly included in the optimization objective under the Maximum A Posterior (MAP) framework. The key insight of the energy-based formulation is to conveniently regularize the motion predictions by customizing the energy functions. In our design, we include two types of energy functions: (1) a frame-wise velocity reconstruction term, and (2) a whole sequence accuracy term. \re{Both terms of (1) and (2) are measured by L2 distance.} Consequently, optimizing the resulting prior losses drives the predictions towards a low-energy configuration to maintain motion coherence and achieve decent prediction accuracy. Furthermore, the introduction of MDN to motion prediction tasks enjoys a considerable advantage in capturing a diverse set of motion hypotheses because of its multimodal nature, and it is difficult for conventional style of regression-based methods, which use a single distribution (e.g., Gaussian), to achieve similar performance. Once trained, our model is able to predict diverse plausible 3D future motions that satisfy the input image. 

In summary, this paper makes the following main contributions:
 \begin{itemize}
\item We propose a novel perspective for addressing the inherent human motion stochasticity arising from the weakly informed conditions (i.e., a single image) by giving diverse plausible predictions, rather than treating it as a simple deterministic regression task. 

\item We model human motion from the image domain to 3D space together with the inner ambiguity under mixture density networks, which has been sparsely treated in 3D human motion prediction tasks, and formulate energy-based priors to improve the prediction performance.

\item We quantitatively and qualitatively report extensive experimental results, as well as ablation studies, on Human3.6M \cite{ionescu2013human3} and MPII \cite{andriluka20142d} datasets, and show the robustness of our method in terms of prediction accuracy and diversity.

 \end{itemize}

\section{Related Work}

\begin{figure*}[ht]
\begin{center}
\includegraphics[width=0.9\linewidth]{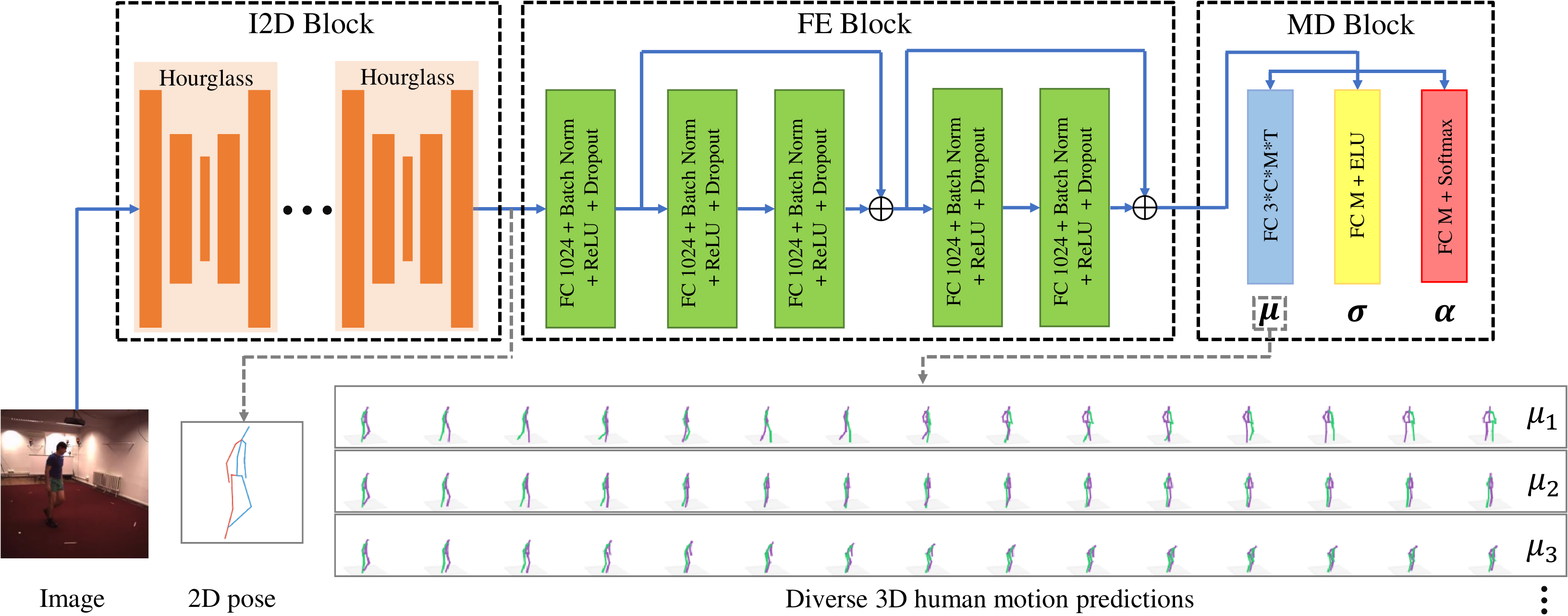}
\end{center}
\caption{Network architecture of our proposed diverse human motion prediction. Blue solid arrows indicate data flow, and the gray dotted arrows denote the output in the corresponding part. FC $n$ refers to a fully-connected layer with output size $n$. The I2D Block is only employed in the test stage. Given a single image as input, diverse 3D human motion predictions are represented with the mean values $\mu_m$ (for $m=1,...,M$) output by the final MD block.}
\label{fig:network}
\end{figure*}

In this section, we first review previous human motion prediction techniques. We then introduce the mixture density networks in a few visual computing tasks, especially regarding human-related modeling. Finally, we briefly review some works on diverse human motion prediction.

\textbf{Human motion prediction.} 
The majority of previous human motion prediction works \cite{fragkiadaki2015recurrent, jain2016structural, butepage2017deep, martinez2017human, gopalakrishnan2019neural, aksan2019structured, wang2019imitation, li2018convolutional, mao2021multi,mao2019learning,mao2020history} rely on \re{the historically observed sequences.} Because human motion is highly stochastic, the most likely future motions can be better captured with the high-dimensional features extracted from the past motions. In \cite{mao2019learning}, the temporal observed sequence is encoded in trajectory space with the Discrete Cosine Transform (DCT). 
Li et al. \cite{li2018convolutional} used a long-term hierarchical convolutional encoder to learn spatial dependencies. The valuable keys in past motions are also discovered with empirical observations. Mao et al. \cite{mao2020history} argued that the history of human motion tends to repeat itself and designed a motion attention module to model the similarity of past and future motions. After extracting the rich information from the past motions, a general style of generating the desired future sequence is to utilize RNN. Besides the training difficulty \cite{pascanu2013difficulty}, \re{RNN generally struggles on long-term prediction scenarios due to the irreversible error accumulation over time \cite{butepage2017deep}}; hence, the prediction performance can degrade greatly with increasing iterations. Despite the efforts on structural improvements to RNN (e.g., three-layer bio-mechanical RNN \cite{du2019bio}, structural RNN with spatio-temporal graph \cite{jain2016structural}), \re{the global smoothness in human motion can still be hard to guarantee.}

The second category predicts human motions under a much weaker condition, e.g., only a single image is given. Human motion prediction in such case is more challenging because a single image contains far less prior information than a motion sequence, leading to more ambiguities. The first attempt to solve this task is by Chao et al. \cite{chao2017forecasting}. In this work, human poses are recurrently regressed in the 2D domain with Long Short Term Memory (LSTM), then the 2D poses are lifted to the 3D domain with an extra 3D converter. Later, \cite{kanazawa2019learning, zhang2019predicting} jointly trained a hallucinator while learning 3D human shapes from videos. The hallucinator aims to hallucinate a very short span of near past and future human motions conditioned on a video clip. Although these works provide some insights into discovering the weakly informed condition, they cannot model the stochasticity imposed by the single image. 

\textbf{Mixture density networks (MDN)}. MDN is first introduced by \cite{bishop1994mixture} to model ill-posed problems by learning the parameters of a mixture distribution. This idea has recently been proven highly impressive in handling human-related multimodal modeling tasks \cite{makansi2019overcoming,prokudin2018deep,choi2018uncertainty}. Makansi et al. \cite{makansi2019overcoming} proposed a Winner-Take-All (WTA) loss to overcome the training instability of MDN for predicting future vehicle location. Choi et al. \cite{choi2018uncertainty} introduced uncertainty-aware learning using MDN to model noisy human behavior in autonomous driving. At the same time, MDN has also been adopted in human pose estimation tasks \cite{varamesh2020mixture, li2019generating, zou2021gmdn, ye2018occlusion}. \cite{varamesh2020mixture} exploited MDN to simultaneously generate a human pose and its location, by endowing each component in the mixture distribution with different meanings. \cite{ye2018occlusion} improved an MDN variant with hierarchical architecture to handle hand joint occlusions. Among these works, Li et al. \cite{li2019generating} shares the most similar motivation to ours in using MDN. They employed MDN to generate multiple possible 3D human poses from the 2D domain to address the ambiguity from view angle and depth. The most distinct point in which our method differs from these works is that we leverage MDN to model human motion, which is more challenging since the increase of the degree of freedom in motion sequence further amplifies ambiguity. Also, we propose a novel perspective by incorporating the energy-based formulation into the MDN framework to further improve the prediction performance, which is not explored in the previous MDN-based approaches. 

\textbf{Diverse human motion prediction.} In many recent works \cite{aliakbarian2019contextually,kania2021trajevae,zhang2021we,yuan2020dlow,barsoum2018hp}, the authors pointed out that even if a past sequence is given, the problem of motion prediction can still be highly ill-posed, and generating a diverse set of probable predictions can be a more reasonable choice. These works share one common motivation that the diversity can be promoted by feeding the deep generative model (e.g., Conditional Variational Auto-Encoder (CVAE) \cite{aliakbarian2019contextually,kania2021trajevae,zhang2021we,yuan2020dlow}, Generative Adversarial Networks (GAN) \cite{barsoum2018hp}) with different sampled noise. In particular, both the encoder and the decoder are under RNN architecture. However, as shown in \cite{yuan2020dlow}, sampling randomly from the latent space is not guaranteed to produce diverse motion samples, and most of them might concentrate on one major mode. Consequently, special subsequent sampling strategies should be supplementarily learned. In addition, none of these works can take the input as a single image.

\section{Method}
Fig. \ref{fig:network} illustrates our network architecture for predicting future human motions from a single image, which consists of an image-to-2D (I2D) block, a feature extraction (FE) block and a mixture density (MD) block.
We first follow the common 2D skeleton extraction policy \cite{martinez2017simple} by adopting the stacked hourglass (SHG) \cite{newell2016stacked} to build the I2D block. \re{The SHG involves repeated hourglass-like encoder-decoder structures and outputs 2D human pose estimations from images.} Specifically, given an image $\mathbf{I}$, the I2D block generates the corresponding 2D human joint detection $\mathbf{x} \in \mathbb{R}^{2C}$, where $C$ is the number of joints in one skeleton. Fed with the 2D poses, the FE block readily extracts the features and maps them into a higher dimensional space. The features then pass through the final MD block to output the parameters of a mixture distribution. The parameters for such distribution are learnable and are trained to achieve a probabilistic mapping from 2D poses to diverse 3D motions under the ground-truth supervision. Note that our training process detours the I2D block and relies on the provided ground-truth 2D annotations of the corresponding images to make full use of the dataset. 

The rest of this section is organized as follows: First of all, we present the overall MDN framework for stochastic human motion prediction (Sec. \ref{sec:sec3.1}). We next describe the optimization strategy for the MDN mappings under the MAP formulation (Sec. \ref{sec:sec3.2}). Then, we detail our energy-based formulation that enables imposing prior losses by customizing the energy function (Sec. \ref{sec:sec3.3}) and explain the Dirichlet conjugate prior that circumvents the degeneration (Sec. \ref{sec:sec3.4}). Eventually, we introduce a training technique to resolve the underflow issue during optimization (Sec. \ref{sec:sec3.5}) and show the qualitative metric we use for evaluation  (Sec. \ref{sec:sec3.6}).

\subsection{Mixture Modeling for Motion Generation}
\label{sec:sec3.1}
Given the 2D human pose $\mathbf{x}$, our goal is to learn a mapping 
function $G(\mathbf{x})$ to generate 3D hypotheses from the unknown multimodal posterior $p(\mathbf{y}|\mathbf{x})$, where $\mathbf{y} \in \mathbb{R}^{3TC}$ indicates the 3D future human motion with the time horizon of $T$ frames. Due to the human motion stochasticity, we propose modeling $p(\mathbf{y}|\mathbf{x})$ under MDN, which can be formulated as
\begin{equation}
\label{eq:eq1}
p(\mathbf{y}|\mathbf{x}) = \sum_{m=1}^M\alpha_m(\mathbf{x}) \phi_m(\mathbf{y}|\mathbf{x}),
\end{equation}
in which $\alpha_m(\mathbf{x}) \in \mathbb{R}$ is the mixture coefficient for the $m$-th component $\phi_m(\mathbf{y}|\mathbf{x})$. \re{Each $\phi_m(\mathbf{y}|\mathbf{x})$ is the probabilistic density for $\mathbf{y}$ conditioned on $\mathbf{x}$ and in this paper is represented by Gaussian distribution:}
\begin{equation}
\label{eq:eq2}
\phi_m(\mathbf{y}|\mathbf{x}) = \frac{1}{{(2 \pi)^{d/2}\sigma^d_{m}(\mathbf{x})}}e^{-\frac{||\mathbf{y}-\mu_m(\mathbf{x})||^2_2}{2\sigma^2_{m}(\mathbf{x})}},
\end{equation}
where $\mu_m(\mathbf{x}) \in \mathbb{R}^{3TC}$ and $\sigma_m(\mathbf{x}) \in \mathbb{R}$ are the mean and standard deviation for $m$-th Gaussian component, respectively. $d=3TC$ \re{refers to} the output dimension of $\mathbf{y}$. $\sigma_m(\mathbf{x})$ is implemented as isotropic for simplicity. Note that the mixture model can also be selected as other forms of distributions (e.g., Laplacian), and we utilize Gaussian mainly because of its convenient mathematical properties. Under such a formulation, $p(\mathbf{y}|\mathbf{x})$ is expressed as a Gaussian mixture, with the parameters $\boldsymbol{\alpha} = \begin{Bmatrix}\alpha_1, ..., \alpha_m, ..., \alpha_M\end{Bmatrix}$, $\boldsymbol{\mu} = \begin{Bmatrix}\mu_1, ..., \mu_m, ..., \mu_M\end{Bmatrix}$, and $\boldsymbol{\sigma} = \begin{Bmatrix}\sigma_1, ..., \sigma_m, ..., \sigma_M\end{Bmatrix}$. Each $\mu_m$ indicates a possible 3D future motion hypothesis with the length of $T$ frames conditioned on $\mathbf{x}$. It is worth mentioning that the conventional motion prediction in a regression fashion can be viewed as a degeneration case of our model by forcing all the Gaussian components to be similar \re{(i.e., $\phi_1 \approx \phi_2 \approx ... \approx\phi_M$)}. The model thus reduces to a single Gaussian, \re{and only captures one possible hypothesis} among potentially diverse modes. We next need to seek the strategy for optimizing the mapping $G(\mathbf{x})$. 


\subsection{Optimization}
\label{sec:sec3.2}
In essence, the target mapping $G(\mathbf{x})$ can be regarded as a triplet mappings of $\Theta(\mathbf{x})=\begin{Bmatrix} \boldsymbol{\alpha}(\mathbf{x}), \boldsymbol{\mu}(\mathbf{x}), \boldsymbol{\sigma}(\mathbf{x}) \end{Bmatrix}$, which are the outputs of the MD block in our network. Particularly, by denoting the network parameters for the mapping $G(\mathbf{x})$ as $\psi$, the whole deep neural network in the function form can be written as
\begin{equation}
\label{eq:eq3}
\Theta(\mathbf{x},\psi) = G(\mathbf{x};\psi).
\end{equation}
Our goal is then switched to finding the optimized parameters $\psi$ that form the ideal mapping $G$. Given $\begin{Bmatrix}\mathbf{Y, X}\end{Bmatrix} = {\begin{Bmatrix}\begin{Bmatrix}\mathbf{y}_k, \mathbf{x}_k\end{Bmatrix} | k=1,...,K\end{Bmatrix}}$ as the training set of ground-truth 3D and 2D pairs, the posterior of $\psi$ can be attained by employing the Maximum A Posterior (MAP) strategy:  
\begin{equation}
\label{eq:eq4}
\begin{aligned}
p(\psi|\mathbf{X},\mathbf{Y}, \Omega) & = \frac{p(\psi|\Omega)p(\mathbf{X}|\psi)p(\mathbf{Y}|\mathbf{X},\psi)}{P(\mathbf{X,Y})} \\ 
&\propto p(\psi, \mathbf{X} | \Omega)p(\mathbf{Y}|\mathbf{X}, \psi)\\
&\propto p(\psi | \mathbf{X}, \Omega)p(\mathbf{Y}|\mathbf{X}, \psi), 
\end{aligned}
\end{equation}
in which $\Omega$ is a newly introduced parameter to characterize the prior knowledge on $\psi$. Furthermore, by assuming the training samples to be i.i.d., we obtain 
\begin{equation}
\label{eq:eq5}
\begin{aligned}
p(\psi|\mathbf{X},\mathbf{Y}, \Omega) &\propto p(\psi | \mathbf{X}, \Omega)\prod_{k=1}^Kp(\mathbf{y}_k|\mathbf{x}_k, \psi) \\
& = p(\psi | \mathbf{X}, \Omega)\prod_{k=1}^K\sum_{m=1}^M\alpha_m(\mathbf{x}_k, \psi) \phi_m(\mathbf{y}_k|\mathbf{x}_k, \psi),
\end{aligned}
\end{equation}
where the parameters for the MDN are rewritten to show dependency on $\psi$ (i.e., $\Theta(\mathbf{x},\psi)=\begin{Bmatrix} \boldsymbol{\alpha}(\mathbf{x}, \psi), \boldsymbol{\mu}(\mathbf{x},\psi), \boldsymbol{\sigma}(\mathbf{x},\psi) \end{Bmatrix}$). 

The optimization of $\psi$ can be thus realized by minimizing the negative log-posterior, which is equivalent to the loss function we use to train our network:
\begin{equation}
\label{eq:eq6}
\begin{aligned}
L &= -\log p(\psi|\mathbf{X},\mathbf{Y}, \Omega) \\
  &\propto -\sum_{k=1}^K \log  p(\mathbf{y}_k|\mathbf{x}_k,\psi) - \log p(\psi|\mathbf{X}, \Omega) \\
  &= 	\underbrace{-\sum_{k=1}^K \log \sum_{m=1}^M\alpha_m(\mathbf{x}_k, \psi) \phi_m(\mathbf{y}_k|\mathbf{x}_k, \psi)}_{L_{MD}} 
  \underbrace{-\log p(\psi|\mathbf{X}, \Omega)}_{L_{pr}} \\
  &= L_{MD} + L_{pr}
\end{aligned}
\end{equation}
\re{It can be observed that the term $L_{MD}$ stands for the likelihood}, which is the general choice of objective for optimizing the MDN. Differently, the MAP formulation is more beneficial as the convenient prior mechanism can play target-specific roles by flexibly designing $\Omega$. Specifically,
$L_{pr}$ can be further evaluated as
\begin{equation}
\label{eq:eq7}
\begin{aligned}
L_{pr} &= -  \log p(\psi|\mathbf{X}, \Omega) \\
&= -  \log p(\psi, \mathbf{X}|\Omega) + \log p(\mathbf{X}|\Omega).
\end{aligned}
\end{equation}
By substituting $\psi, \mathbf{X}$
with its function form $\Theta(\psi, \mathbf{X})$, and dropping the $\psi$-independent $\log (\mathbf{X}|\Omega)$ term, we obtain
\begin{equation}
\label{eq:eq8}
\begin{aligned}
L_{pr} &\propto -  \log p(\Theta(\psi, \mathbf{X}) | \Omega) \\
&= -\log p(\boldsymbol{\mu}(\psi,\mathbf{X} | \omega_1))  -\log p(\boldsymbol{\alpha}(\psi,\mathbf{X} | \omega_2)) \\
&\quad-\log p(\boldsymbol{\sigma}(\psi,\mathbf{X} | \omega_3)),
\end{aligned}
\end{equation}
where $\boldsymbol{\mu}$, $\boldsymbol{\alpha}$ and $\boldsymbol{\sigma}$ are assumed to be mutually independent. \re{We further divide $\Omega$ into $\begin{Bmatrix}\omega_1, \omega_2 ,\omega_3\end{Bmatrix}$
to represent prior knowledge on different parameters of the MDN. Next, we need to determine the design policy of $\Omega$ to fulfill different demands in the task of human motion prediction.}


\subsection{Energy-based Prior Loss} 
\label{sec:sec3.3}
\re{To enable each mean of the MDN (i.e., $\mu_m$) to represent the whole motion sequence}, the prior $\omega_1$ should be elaborately defined. Although our model does not suffer from the error accumulation issue in RNN, it also does not benefit from the feature passed from the past outputs that drives the generated sequence to be coherent. \re{Therefore, the prior $\omega_1$ needs to be capable of maintaining coherence.} We draw inspiration from \cite{yuan2020dlow} by leveraging the energy-based formulation over the learnable priors. The core advantage of such formulation lies in the design flexibility of the underlying energy functions, with which we can impose versatile constraints during optimization. Under the energy-based formulation, we derive an energy function $E_v$ by defining
\begin{equation}
\label{eq:eq9}
p(\boldsymbol{\mu}) = e^{-E_v(\boldsymbol{\mu}) / S},
\end{equation}
in which $S$ is a normalizing constant, and can be dropped in the optimization. The notation of  $p(\boldsymbol{\mu}(\psi,\mathbf{X} | \omega_1)$ is simplified as $p(\boldsymbol{\mu})$ for brevity. The corresponding loss with respect to $\boldsymbol{\mu}$ in Eq. \ref{eq:eq8} is thus given by
\begin{equation}
\label{eq:eq10}
\log p(\boldsymbol{\mu}) = -E_v(\boldsymbol{\mu}).
\end{equation}
We use frame-wise velocity between the ground-truth and the predicted sequence to measure the degree of motion consistency and introduce $f^{t}_{m}(\mathbf{x}_k) \in \mathbb{R}^{3C}$ to denote the pose at the $t$-th frame in the $m$-th motion hypothesis $\mu_m(\mathbf{x}_k)$. $f^{t}_{m}(\mathbf{x}_k)$ can also be considered a direct mapping that outputs a 3D skeleton given the $k$-th training sample $\mathbf{x}_k$, and $f^{1}_{m}(\mathbf{x})$ refers to the 3D pose of the 2D joints $\mathbf{x}_k$ for $m=1 ..., M$.
Given training input $\mathbf{x}_k$, the velocity vector for the $m$-th motion hypothesis can be computed as $\textbf{v}^{k}_m = [f^2_{m}(\mathbf{x}_k),...,f^T_{m}(\mathbf{x}_k)] - [f^1_{m}(\mathbf{x}_k),...,f^{T-1}_{m}(\mathbf{x}_k)]$. We hence form the energy function $E_v$ based on the pairwise Euclidean distance $\mathcal{D}$ between predicted velocity $\textbf{v}^{k}_m$ and the corresponding ground-truth velocity $\textbf{v}^{k}$,
\begin{equation}
\label{eq:eq11}
E_v = \frac{1}{M(T-1)}\sum^{K}_{k=1}\sum^{M}_{m=1} (\mathcal{D}(\textbf{v}^{k}_m, \textbf{v}^{k}))^2.
\end{equation}
Eventually, minimizing $E_v$ guides the generated prediction toward a low-energy configuration such that coherent motions can be pursued.

Although our model produces diverse motion predictions,
we also expect a decent prediction accuracy. To this end, we modify the energy term in Eq. $\ref{eq:eq9}$ by adding an accuracy term 
\begin{equation}
\label{eq:eq12}
E_a = \sum^{K}_{k=1}\underset{m}{\mathrm{min}}(\mathcal{D}(\mu_m(\mathbf{x}_k), \mathbf{y}_k))^2,
\end{equation}
to hopefully ensure that at least one of the generated predictions $\mu_m$ can precisely characterize the ground-truth $\mathbf{y}_{k}$.


\subsection{Dirichlet Conjugate Prior Loss}
\label{sec:sec3.4}
The mixture coefficient $\boldsymbol{\alpha}$ of the MDN can be regarded as the probability of the prediction $\mathbf{y}$ being generated from the $m$-th Gaussian component. As an auxiliary degeneration-preventing training assistant, we adopt the strategy in \cite{li2019generating, gu2021cgmvae} by applying a Dirichlet conjugate loss as a regularization term to preserve the multimodality of the MDN. The Dirichlet conjugate prior over mixture coefficient $\boldsymbol{\alpha}$ is computed by:
\begin{equation}
\label{eq:eq13}
\begin{aligned}
p(\alpha_1, ..., \alpha_M | \Lambda) &= Dir_{[\alpha_1, ..., \alpha_M]}[\lambda_1, ..., \lambda_M]\\
&=\frac{\Gamma[\sum_{m=1}^M\lambda_m]}{\prod_{m=1}^{M}\Gamma[\lambda_m]}\prod_{m=1}^{M}\alpha_m^{(\lambda_m-1)},
\end{aligned}
\end{equation}
in which $\Gamma[\cdot]$ denotes the Gamma function, and $\Lambda = \begin{Bmatrix}\lambda_1, ..., \lambda_M\end{Bmatrix}$ represents the parameters for Dirichlet distribution. 

The final Dirichlet conjugate loss is thus defined by 
\begin{equation}
\label{eq:eq14}
\begin{aligned}
L_d &= -\log(p(\boldsymbol{\alpha})|\Lambda)\\
& = -\sum_{k=1}^{K}\sum_{m=1}^{M}(\lambda_m-1)\log \alpha_m(\mathbf{x}_k), 
\end{aligned}
\end{equation}
in which the $\frac{\Gamma[\sum_{m=1}^M\lambda_m]}{\prod_{m=1}^{M}\Gamma[\lambda_m]}$ part is dropped since it is independent of $\alpha_m$. Generally, $\lambda_m = 1$ for $m=1,...,M$ implies that there is no prior knowledge over $\bm{\alpha}$. We set $\lambda_1 = ... = \lambda_M = 2$ as suggested in \cite{li2019generating}. Even though our method does not trigger noticeable degeneration, \re{we have observed that including $L_d$ well refines prediction performance (see Sec. \ref{sec:Ablation Studies})}.

\subsection{Training Technique}
\label{sec:sec3.5}
Up to this point, we have explained the proposed diverse motion prediction method, and introduced different prior terms $\Omega$ over $\boldsymbol{\mu}$ and $\boldsymbol{\alpha}$. The remaining prior for $\boldsymbol{\sigma}$ is simply assumed to be uniform. By substituting Eq. \ref{eq:eq6} for Eq. \ref{eq:eq11}, Eq. \ref{eq:eq12}, and Eq. \ref{eq:eq13}, the overall loss we use to train our model is given by
\begin{equation}
\label{eq:eq15}
L = L_{MD} + \mathbf{w}_v * L_v + \mathbf{w}_a *L_a + \mathbf{w}_d *L_d, 
\end{equation}
where
\begin{equation}
\label{eq:eq16}
\begin{aligned}
L_{MD} &= -\sum_{k=1}^K \log \sum_{m=1}^M\alpha_m(\mathbf{x}_k, \psi) \phi_m(\mathbf{y}_k|\mathbf{x}_k, \psi),  \\
L_v &= E_v,  \\
L_a &= E_a,
\end{aligned}
\end{equation}
and $\mathbf{w}_v$, $\mathbf{w}_a$ and $\mathbf{w}_d$ are the weights for the corresponding loss, respectively. 

We next introduce a training technique in terms of $L_{MD}$. 
Formally, given an arbitrary training sample $\mathbf{x}$, $L_{MD}$ can be calculated as 
\begin{equation}
\label{eq:eq17}
L_{MD} = -\log \sum_{m=1}^M\frac{\alpha_m(\mathbf{x})}{(2\pi)^{d/2}\sigma_m^2(\mathbf{x})}e^{-\frac{||\mathbf{y}-\mu_m(\mathbf{x})||_2^2}{2\sigma^2_{m}(\mathbf{x})}}.
\end{equation}
In the training process, the right-most exponential term in Eq. \ref{eq:eq17} is highly likely to be excessively small, and the outside logarithm operation can induce the underflow issue. To mitigate the resulting negative influence and ease training, we leverage the log-sum-exp trick \cite{Guillaumes2017MixtureDN} that separates the maximum within the exponential to the outside:
\begin{equation}
\label{eq:eq18}
\log\sum_{j=1}^{{J}}e^{q_j} = \underset{j}{\mathrm{max}}(q_j) + \log\sum_{j=1}^{{J}}e^{(q_j - \underset{j}{\mathrm{max}}(q_j))}.
\end{equation}
By extending the exponential inside the logarithm:
\begin{equation}
\label{eq:eq19}
L_{MD} = -\log \sum_{m=1}^M e^ {\log \alpha_m(\mathbf{x}) - \frac{d}{2}\log 2\pi \sigma^2_m(\mathbf{x}) -\frac{||\mathbf{y}-\mu_m(\mathbf{x})||_2^2}{2\sigma^2_{m}(\mathbf{x})}},
\end{equation}
the scheme in Eq. \ref{eq:eq18} can then be easily applied. 

\subsection{Evaluation Methods}
\label{sec:sec3.6}
Our model is able to predict multimodal human motion once trained. To demonstrate the effectiveness of our method, we adopt the following two metrics for quantitative evaluation: (i) For prediction accuracy, we measure the Mean Per Joint Position Error (MPJPE) metric in millimeters between the ground-truth and the predicted 3D motions; (ii) For prediction diversity, we measure the Average Pairwise Distance (APD) metric \cite{yuan2020dlow}, which is calculated as $\frac{1}{M(M-1)}\sum_{m=1}^M\sum_{m^{\prime}=1, m^{\prime}\neq m}^M||\mu_{m}-\mu_{m^{\prime}}||_2$. See Sec \ref{sec:sec4.1} and \ref{sec:Ablation Studies} for detailed quantitative analysis.

\section{Experiment}
\noindent	\textbf{Datasets.} We assess our method using the following two human motion/pose estimation benchmarks: Human3.6M \cite{ionescu2013human3} and MPII \cite{andriluka20142d}. Human3.6M is to date the largest and most widely used dataset, which provides accurate 2D and 3D joint annotations on videos recorded by the Vicon Mocap system. It contains 15 daily activity scenarios performed by seven professional actors under four camera views. We conduct both qualitative and quantitative evaluations on this dataset. MPII is a more challenging benchmark for 2D human pose estimation because of various types of background and severe occlusion collected from YouTube videos. We only qualitatively evaluate our method on this dataset, as 3D annotations are not provided. 

\begin{table*}[]
\centering
\caption
{Quantitative results of MPJPE on the Human3.6M for future motion prediction. The best results are in bold. }
\label{tab:tab1}
\scalebox{0.7}{
\begin{tabular}{lcccccccccccccccc}
\hline
{\color[HTML]{333333} {\ul }} & Direct.        & Discuss         & Eating         & Greet           & Phone          & Smoke           & Pose           & Purch.         & Sitting         & SittingD.       & Smoke          & Wait            & Walk            & WalkD.         & WalkT.         & Avg             \\ \hline
(i) Martinez \cite{martinez2017simple}           & 115.47         & 148.12          & 100.65         & 146.47          & 108.59         & 172.54          & 126.64         & 146.61         & 124.38          & 159.79          & 120.69         & 138.42          & 166.02          & 103.55         & 99.89          & 131.85          \\
(ii) Martinez \cite{martinez2017human}             & 114.49         & 146.57          & 115.15         & 151.77          & 112.15         & 153.71          & 130.21         & 148.70         & 119.20          & 162.89          & 120.36         & 139.23          & 181.25          & 159.06         & 163.19         & 141.16          \\
\multicolumn{1}{c}{Ours}      & \textbf{88.60} & \textbf{106.05} & \textbf{84.20} & \textbf{110.06} & \textbf{90.97} & \textbf{135.81} & \textbf{98.05} & \textbf{94.19} & \textbf{101.88} & \textbf{132.66} & \textbf{98.30} & \textbf{103.95} & \textbf{121.18} & \textbf{91.22} & \textbf{85.83} & \textbf{102.86} \\ \hline
\end{tabular}
}
\end{table*}

\begin{table*}[]
\centering
\caption
{Evaluation in terms of MPJPE on the regularizing power of $L_{d}$, $L_{v}$ and $L_{a}$. The best results are in bold. The final row refers to our model with all three types of prior losses involved.}
\label{tab:tab2}
\scalebox{0.67}{
\begin{tabular}{lcccccccccccccccc}
\hline
{\ul }              & Direct.        & Discuss         & Eating         & Greet           & Phone          & Smoke           & Pose           & Purch.         & Sitting         & SittingD.       & Smoke          & Wait            & Walk            & WalkD.         & WalkT.         & Avg             \\ \hline
Ours ($L_{MD}$)       & 92.09          & 112.62          & 87.97          & 116.79          & 94.25          & 143.68          & 101.83         & 100.98         & 106.14          & 138.73          & 101.98         & 109.54          & 127.32          & 94.98          & 91.38          & 108.02          \\
Ours ($L_{MD}$, $L_d$)       & 91.36          & 109.80          & 87.19          & 114.95          & 93.31          & 142.13          & 102.84         & 96.89          & 104.23          & 136.53          & 100.79         & 107.78          & 125.91          & 95.91          & 91.30          & 106.73          \\
Ours ($L_{MD}$, $L_d$, $L_v$)   & \textbf{87.25}          & 106.38          & 85.40          & 110.97          & 91.82          & 138.52          & \textbf{97.60}          & 94.61          & 102.06          & \textbf{132.28}          & 98.46          & 104.65          & 122.79          & 93.81          & 89.21          & 103.72          \\
Ours ($L_{MD}$, $L_d$, $L_v$, $L_a$) & 88.60 & \textbf{106.05} & \textbf{84.20} & \textbf{110.06} & \textbf{90.97} & \textbf{135.81} & 98.05 & \textbf{94.19} & \textbf{101.88} & 132.66 & \textbf{98.30} & \textbf{103.95} & \textbf{121.18} & \textbf{91.22} & \textbf{85.83} & \textbf{102.86} \\ \hline
\end{tabular}}
\end{table*}

\noindent	\textbf{Network and implementation details.} 
Similar to \cite{zhao2019semantic, martinez2017simple}, the I2D block is pretrained on MPII and fined-tuned with Human3.6M to directly estimate 2D poses from images in the test stage. To completely show the power of our MDN modeling, we simply adopt the architecture in \cite{martinez2017simple} as the backbone to construct the FE block. As shown in Fig. \ref{fig:network}, the FE block takes as input a 2D skeleton and lifts it to a 1024 dimensional feature space as pre-processing. The features are then passed into a two-layer residual module, in which batch normalization, dropout (with a rate of 0.5), and ReLU activation are involved after each fully connected (FC) layer. The residual module is repeated for two rounds to deepen the structure. The output of the FE block is fed into the proposed MD block to produce mixture distribution parameters. We use ELU \cite{clevert2015fast} as the activation for $\boldsymbol{\sigma}$. We add a softmax function after the FC layer for $\boldsymbol{\alpha}$ to constrain it with $\sum^M_{m=1}\alpha_m = 1$. No further treatment besides the FC layer upon $\boldsymbol{\mu}$ is applied. 

Our model is trained for 100 epochs on RTX 3090, and the ADAM optimizer \cite{kingma2014adam} is utilized based on an exponentially decaying learning rate with the initial value of 1e-4. The weights $(\mathbf{w}_v$, $\mathbf{w}_a$, $\mathbf{w}_d)$ are empirically set to (0.1, 0.05, 1.0). To avoid training failure in which the $L_{MD}$ deteriorates to NaN, we clip the value of $\sigma_m$ to [1e-5, 1e5], and $\alpha_m$ to [1e-8, 1] for $m=1,...,M$, respectively. For all experiments, we predict the future 15 frames (3 seconds) based on the single input image. We also include a constraining loss to match the first frames of all predictions. To be consistent with the previous literature \cite{martinez2017simple,zhou2017towards,sun2017compositional}, we use S1, S5, S6, S7 and S8 for training, and test on S9 and S11, which is commonly referred to as Protocol $\#$1 \cite{zhao2019semantic}.

\vskip\baselineskip

\noindent	\textbf{Data pre-processing.} 
Following the settings from previous works \cite{wandt2019repnet, zhou2017towards}, we align every 3D pose in the Human3.6M dataset with the provided ground-truth 2D joint locations by applying the camera calibration transformation. The transformation, including a scale, rotation, and translation, is attained from Procrustes analysis on the hip and shoulder joints. We then zero-center both the 2D and 3D poses around the root joint (pelvis). All the sequences are down-sampled to five frames per second to reduce redundancy.

\subsection{Quantitative Evaluation}
\label{sec:sec4.1}
We first report the quantitative results in terms of prediction accuracy on Human3.6M. We follow previous works \cite{yuan2020dlow,li2019generating} by measuring the MPJPE between the ground-truth motion and the best generated motion prediction (i.e., the closest one to the ground-truth). \re{Because human motion prediction from a single image has been rarely studied to date} \footnote{As the motion prediction module in \cite{chao2017forecasting} mainly learns 2D poses and does not involve 3D supervision, we can hardly make a fair comparison as our method is trained with 3D data.}, we devise two original baselines for comparison: (i) We re-train a 3D pose estimation approach \cite{martinez2017simple}, which also takes a single image as input, by stretching the output to be a 3D motion sequence and use the ground-truth motion sequence $\mathbf{Y}$ as supervision; (ii) We re-train a RNN-based 3D sequence-to-sequence motion prediction technique \cite{martinez2017human} only with the human pose in the last frame of observed sequence as input, making the prediction performs in the manner of 3D pose-to-sequence mapping. For both (i) and (ii), we make all the outputs represent a motion sequence and attempt to at least ensure that the inputs are all single. The results are summarized in Tab. \ref{tab:tab1}. \textit{Note that it is, however, not a fair comparison because these two methods are not originally designed for predicting motions from a single image as ours, and the comparative results are only aimed to show the superiority of the MDN modeling in better capturing the correct motion. }  

It can be verified from Tab. \ref{tab:tab1} that our method outperforms \cite{martinez2017simple, martinez2017human} by a large margin in MPJPE, and both of the baselines encounter the bottleneck in capturing the accurate future motions from a weakly-informed single input (i.e., single image for \cite{martinez2017simple} and single 3D pose for \cite{martinez2017human}). Moreover, deterministic systems can be confused by the structurally similar 2D poses followed by distinct future motions, which can be viewed as the main reason for the low MPJPE reflected in Tab. \ref{tab:tab1}.
On the contrary, the multimodality of the MDN demonstrates the effectiveness in accurately predicting motions from a single image.

\begin{figure*}[t]
\begin{center}
\includegraphics[width=0.92\linewidth]{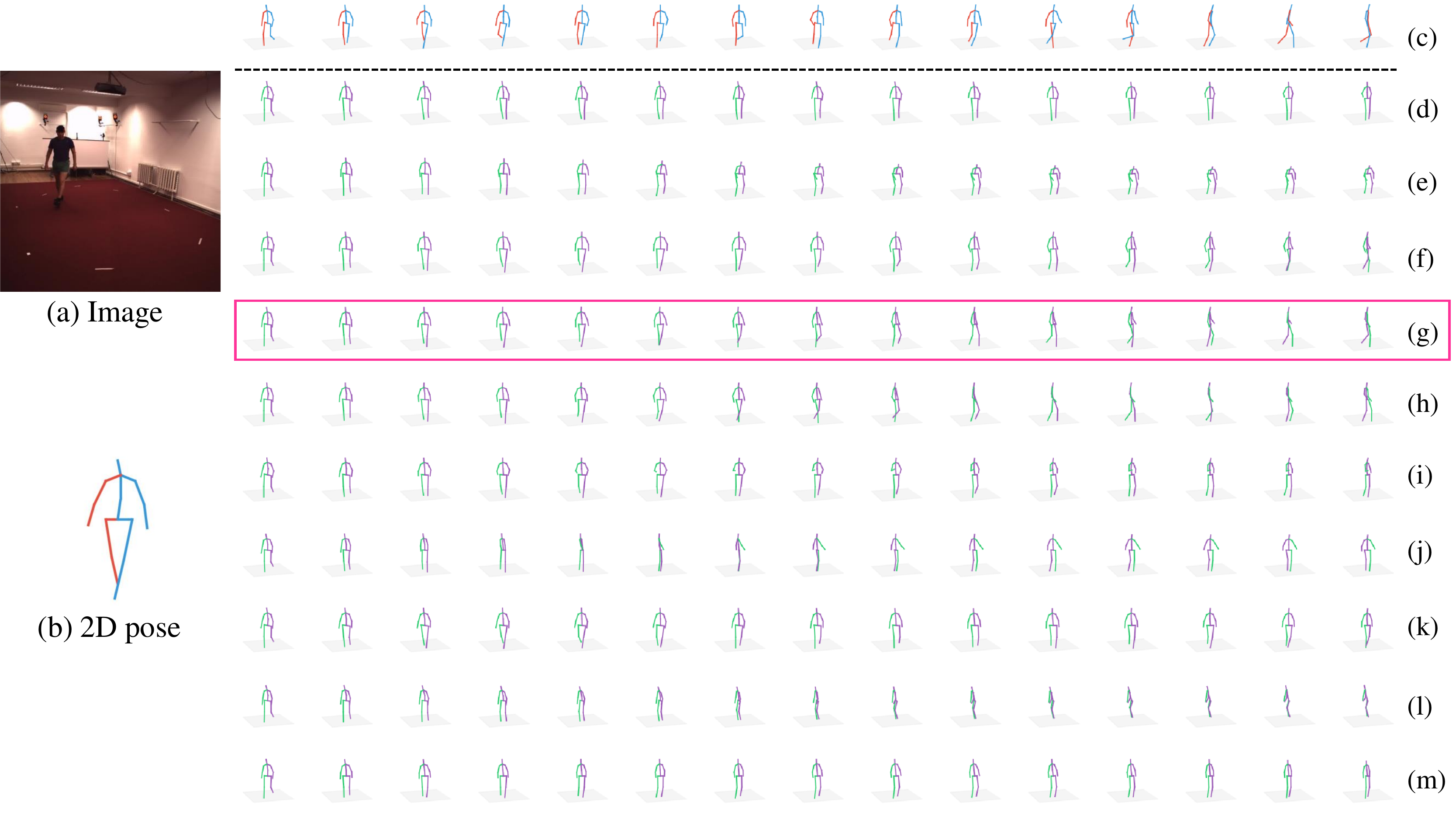}
\end{center}
\caption{Example of diverse future motion prediction results on action ``Walking'' in Human3.6M. (a), (b) are the input image and the corresponding 2D human pose estimated by I2D block, respectively. (c) refers to the ground-truth future motion. (d) $\sim$ (m) represent 10 plausible future motion hypotheses. The highlighted motion (g) with the magenta box denotes the closest motion to the ground-truth (c).}
\label{fig:res1}
\end{figure*}

\begin{figure*}[]
\begin{center}
\includegraphics[width=0.92\linewidth]{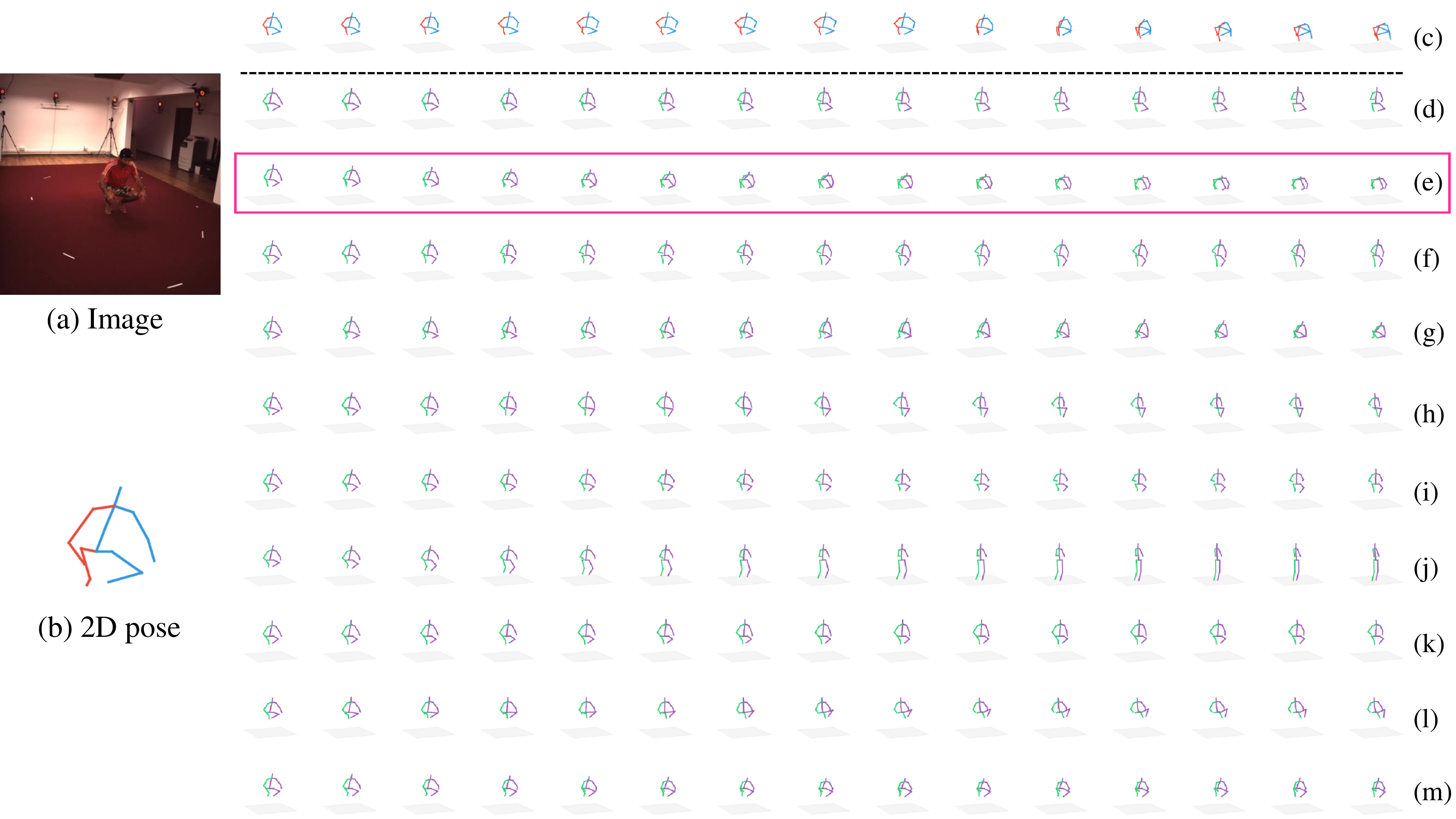}
\end{center}
\caption{ Example of diverse future motion prediction results ((d) $\sim$ (m)) on image (a) with action ``SittingDown'' in Human3.6M. The highlighted motion (e) with the magenta box denotes the closest motion to the ground-truth (c).}
\label{fig:res2}
\end{figure*}

\begin{figure*}[]
\begin{center}
\includegraphics[width=0.92\linewidth]{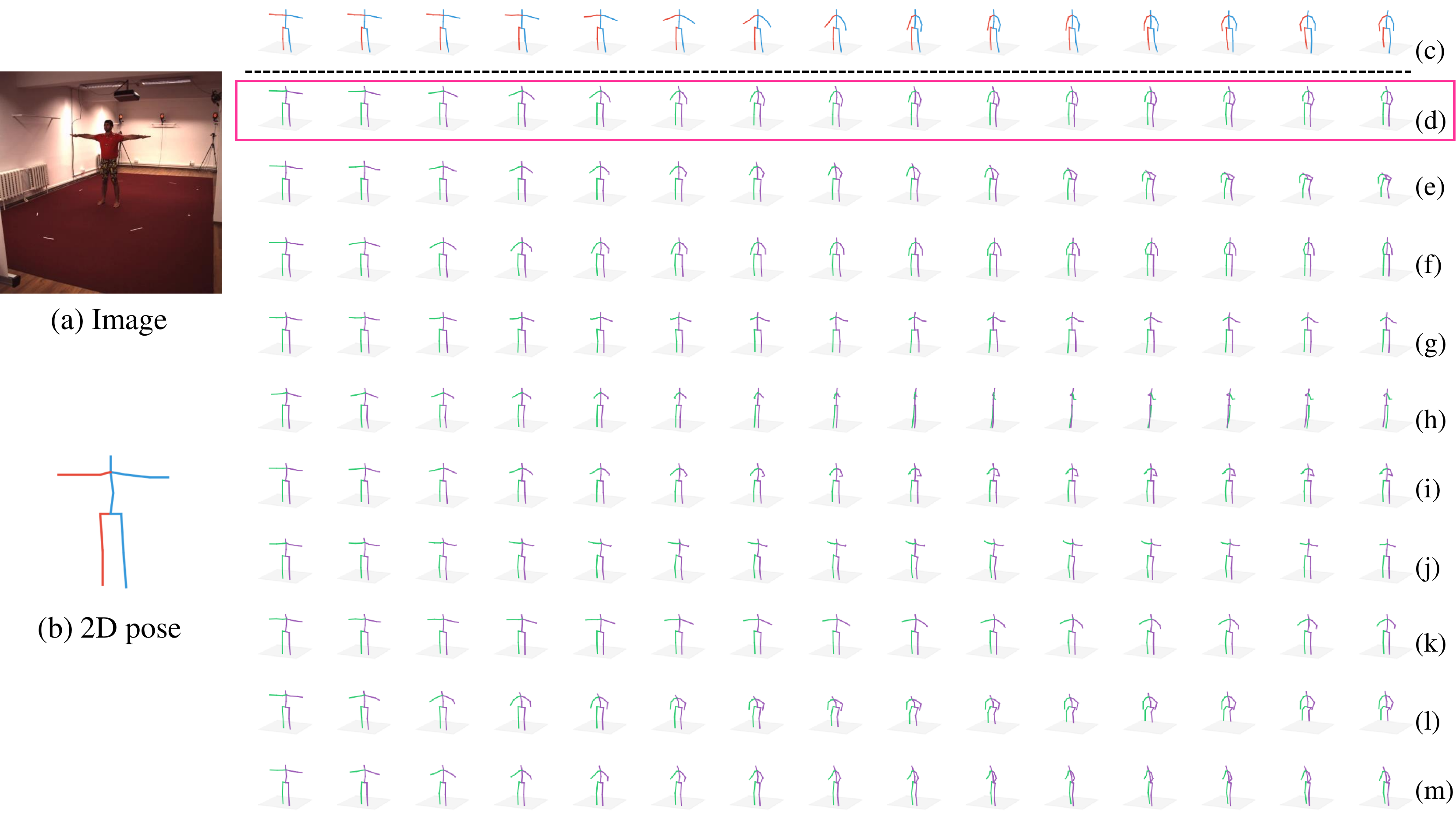}
\end{center}
\caption{ Example of diverse future motion prediction results ((d) $\sim$ (m)) on image (a) with action ``Greeting'' in Human3.6M. The highlighted motion (d) with the magenta box denotes the closest motion to the ground-truth (c).}
\label{fig:res3}
\end{figure*}

\begin{figure*}[]
\begin{center}
\includegraphics[width=0.92\linewidth]{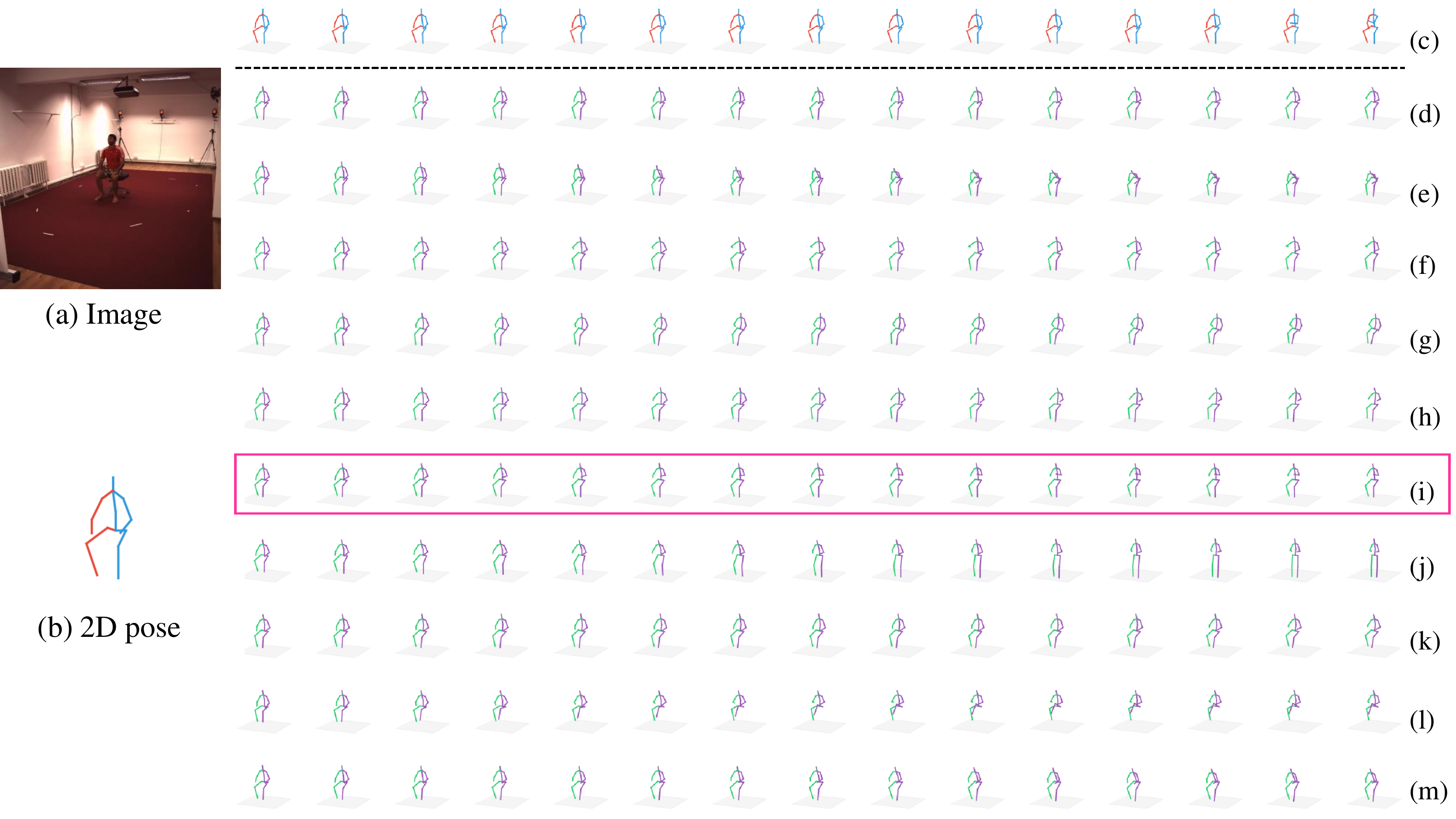}
\end{center}
\caption{ Example of diverse future motion prediction results ((d) $\sim$ (m)) on image (a) with action ``Sitting'' in Human3.6M. The highlighted motion (i) with the magenta box denotes the closest motion to the ground-truth (c).}
\label{fig:res4}
\end{figure*}

\subsection{Qualitative Evaluation}
We hereby show some qualitative human motion prediction results on Human3.6M in Fig. \ref{fig:res1} $\sim$ Fig. \ref{fig:res4}. It can be found that our method can generate visually accurate (or fairly close) predictions among diverse hypotheses for both standing and sitting actions. Also, we observe that action with more ambiguity in the image results in more variations among future motion hypotheses. For example, the half-squat pose in Fig. \ref{fig:res2} can naturally evolve to many possibilities such as squatting deeper (Fig. \ref{fig:res2}(g,e)), rising up (Fig. \ref{fig:res2}(f,j)), sitting down (Fig. \ref{fig:res2}(l,h)), keeping still (Fig. \ref{fig:res2}(m)), etc. Similar results can be seen in Fig. \ref{fig:res3}, where the following hand motions can be lowered (Fig. \ref{fig:res3}(d,e,f,i,k,l,m)) or kept horizontally raised (Fig. \ref{fig:res3}(g,h,j)). In contrast, once the given condition is less ambiguous, like the walking pose in Fig. \ref{fig:res1}, most outputs by our model can be standing-based motions. We suspect the reason could be that the large proportion of standing poses in Human3.6M gives rise to the dataset bias toward standing motions. Nevertheless, even in this case, we can still notice that the model attempts to yield diverse predictions by varying walking direction or adjusting walking velocity. We also show a challenging case in terms of unexpected change in human motions in Fig. \ref{fig:res4}. In Fig. \ref{fig:res4}(c), the last two frames suddenly change while the first 13 frames almost remain still. None of the 10 motion hypotheses generated by our model gives the correct changing time of arm movement in the 14th frame. Exposed to such a huge stochasticity, our model comprises to give several plausible motion hypotheses (Fig. \ref{fig:res4}(d,f,h,i)) to cover lots of probable motion combinations with elbows and wrists. From the above analysis, \re{we can confirm that our proposed MDN in motion prediction tasks} possesses the strength to disambiguate the uncertainty imposed by the single input image and produces visually natural movements that satisfy the given condition.

\begin{figure}[t]
\begin{center}
\includegraphics[width=1.0\linewidth]{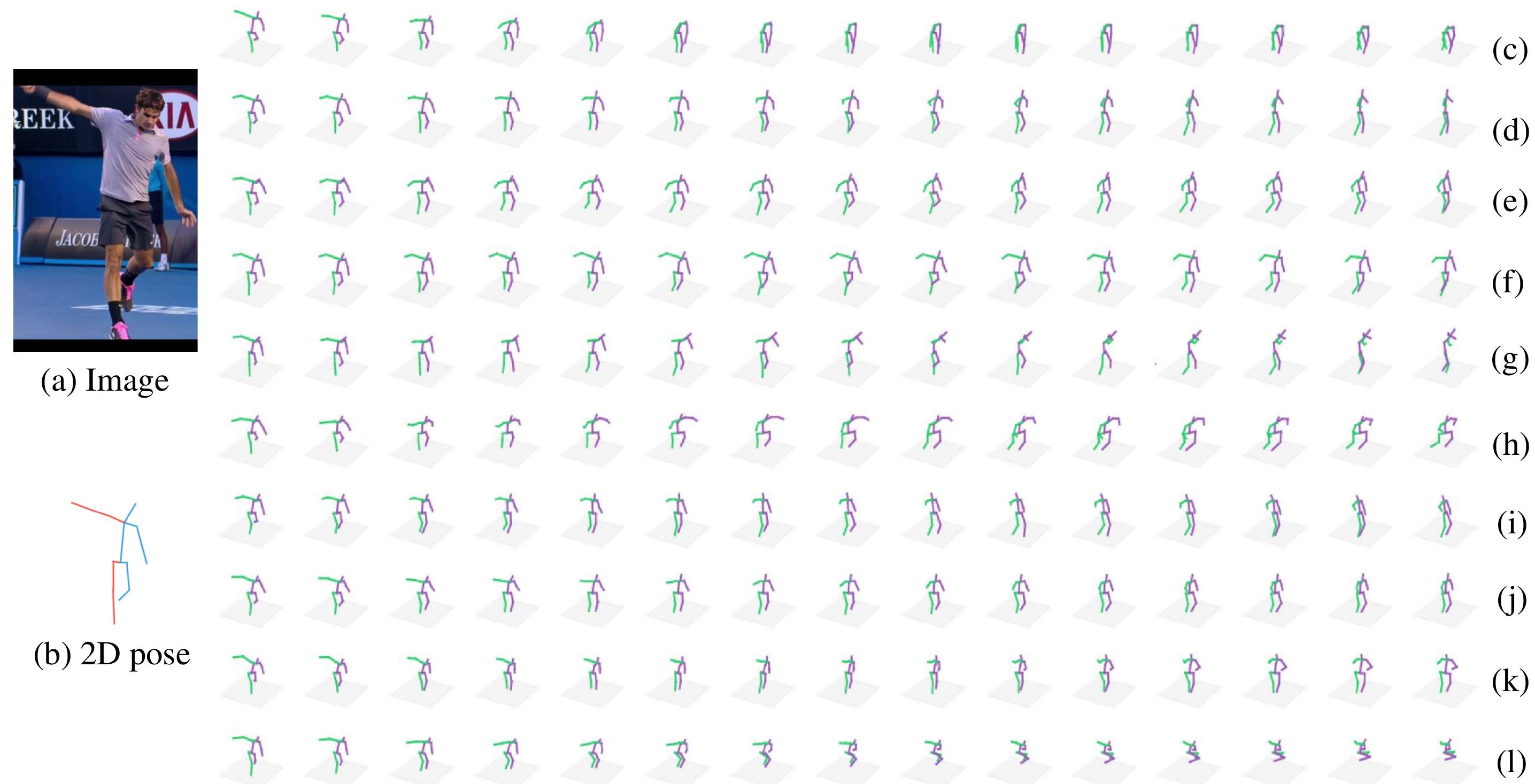}
\end{center}
\caption{Example of diverse future motion prediction results on image ``Man playing tennis'' in MPII. (a), (b) are the input image and the corresponding 2D human pose estimated by I2D block, respectively. (c) $\sim$ (l) represent 10 plausible future motion hypotheses.}
\label{fig:res_MPII_1}
\end{figure}

\begin{figure}[t]
\begin{center}
\includegraphics[width=1.0\linewidth]{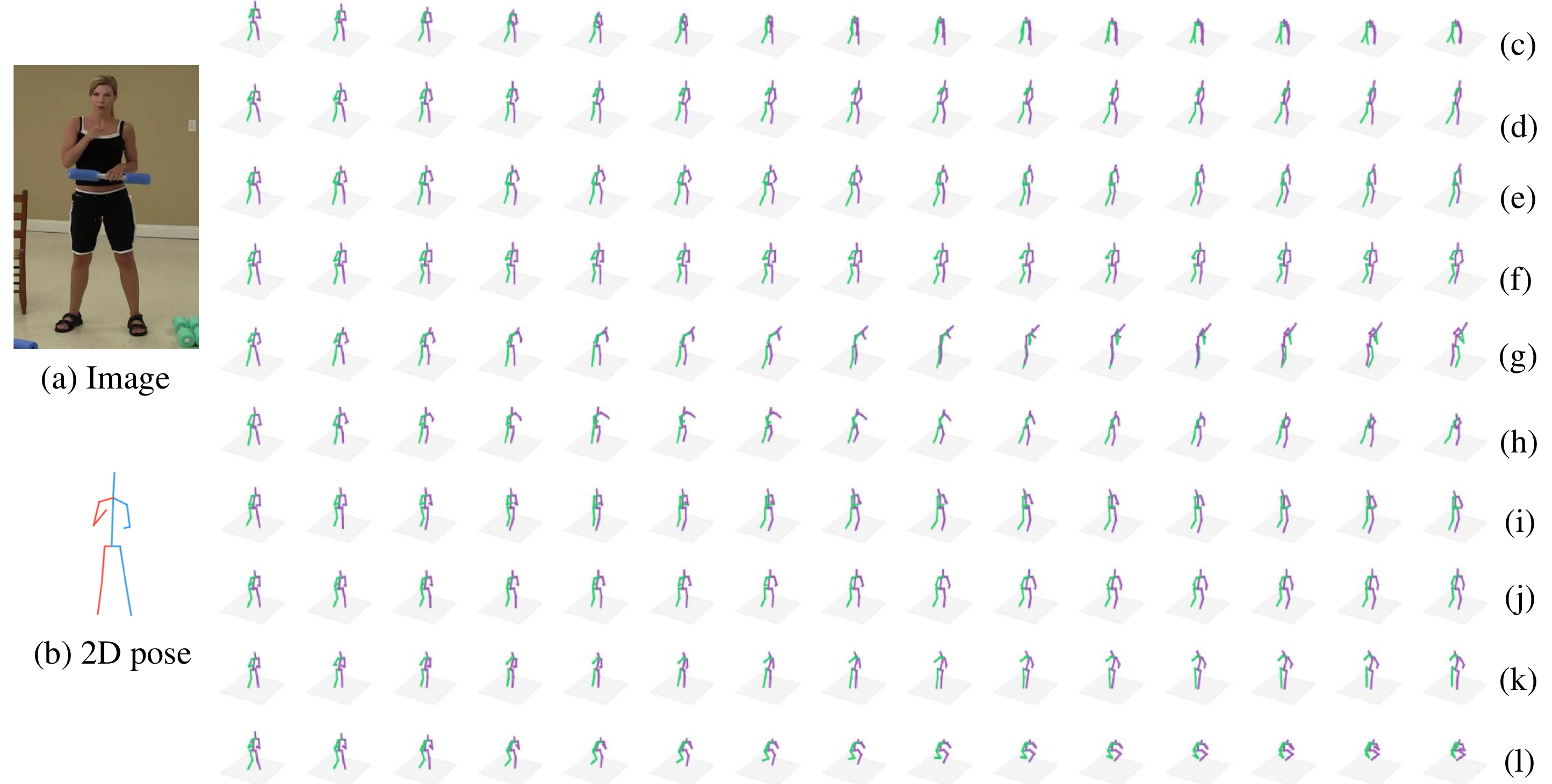}
\end{center}
\caption{Example of diverse future motion prediction results ((c) $\sim$ (l)) on the input image ``Woman standing'' (a) in MPII.}
\label{fig:res_MPII_3}
\end{figure}

We next investigate the generalization capability of our approach by using our trained model on Human3.6M to directly examine the prediction power on MPII. Challenges in transferring our model to MPII mainly lie in: (i) MPII widely includes images from both indoor and outdoor environments, whereas all the scenes in Human3.6M are recorded indoors within a fixed space; (ii) MPII contains various types of 2D poses compared with Human3.6M, with more potential possibilities in the future motions. As illustrated in Fig. \ref{fig:res_MPII_1} $\sim$ Fig. \ref{fig:res_MPII_3}, we can see that our method generally produces reasonable future motions on the given image. It is interesting to point out that in Fig. \ref{fig:res_MPII_1}, our model generates visually convincing future motions that appear similar to the tennis player hitting back the tennis ball (Fig. \ref{fig:res_MPII_1}(g)), even if such sports-related scenes do not exist in the training data. Consequently, the qualitative results in Fig. \ref{fig:res_MPII_1} $\sim$ Fig. \ref{fig:res_MPII_3} demonstrate the robust generalization of our model in handling different types of datasets.

\begin{figure}[t]
\begin{center}
\includegraphics[width=1.0\linewidth]{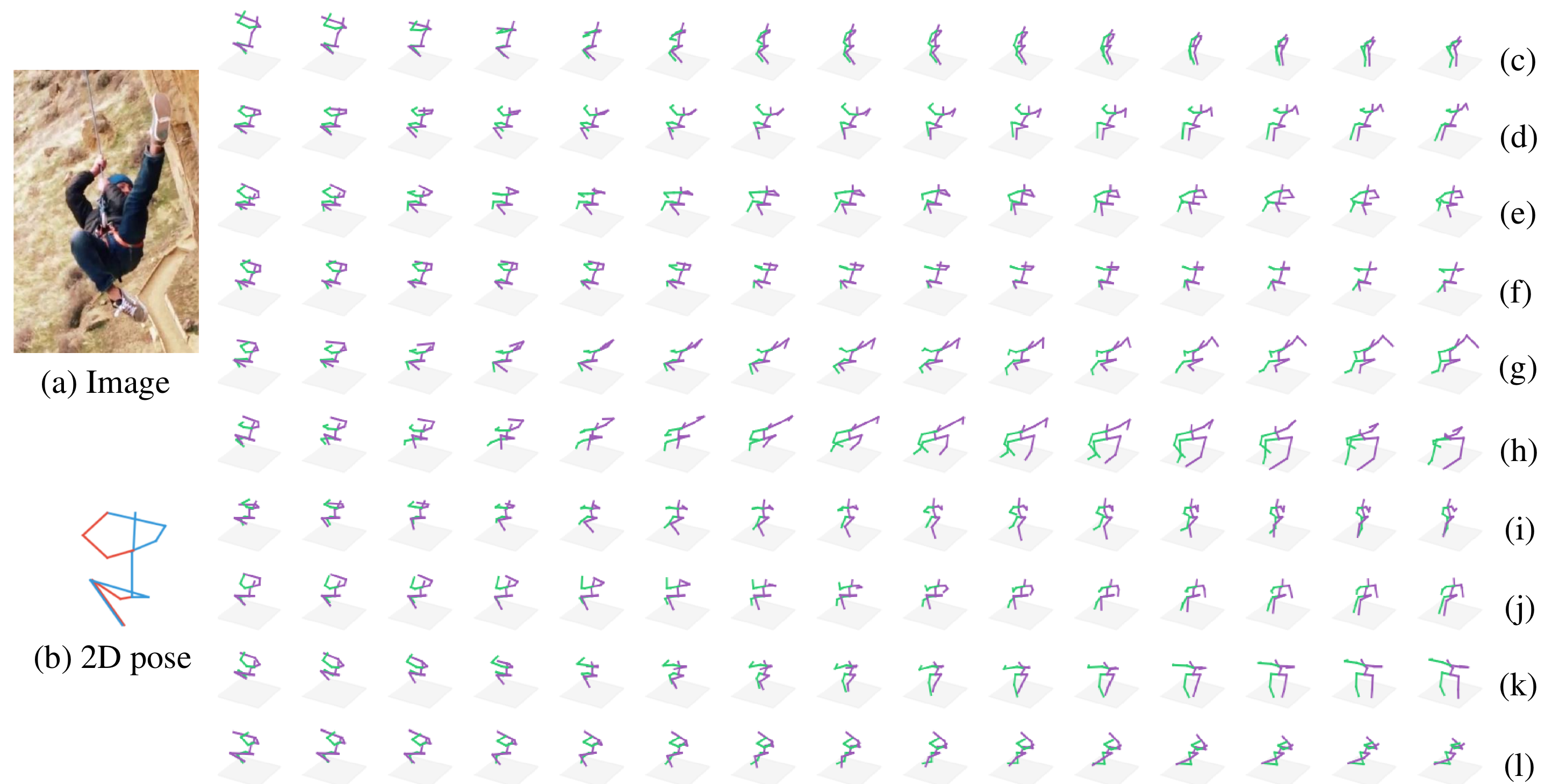}
\end{center}
\caption{A failure case on the input image ``Woman climbing'' (a) in MPII because of wrong 2D joints estimation (b).}
\label{fig:res_MPII_fail}
\end{figure}

We also show a failure case of our method in Fig. \ref{fig:res_MPII_fail}, \re{in which some of the generated motion hypotheses (Fig. \ref{fig:res_MPII_fail}(g,h)) seem to violate the kinematics of human motion.} Also, the first frames (i.e., the corresponding 3D pose for Fig. \ref{fig:res_MPII_fail}(b)) of all the generated motion hypotheses are visually inconsistent with the image (Fig. \ref{fig:res_MPII_fail}(a)).
The reason can be concluded to the significantly wrong 2D pose returned by the I2D block, and our model therefore cannot recover motions correctly from the wrong input.


\begin{figure}[t]
\begin{center}
\includegraphics[width=1.0\linewidth]{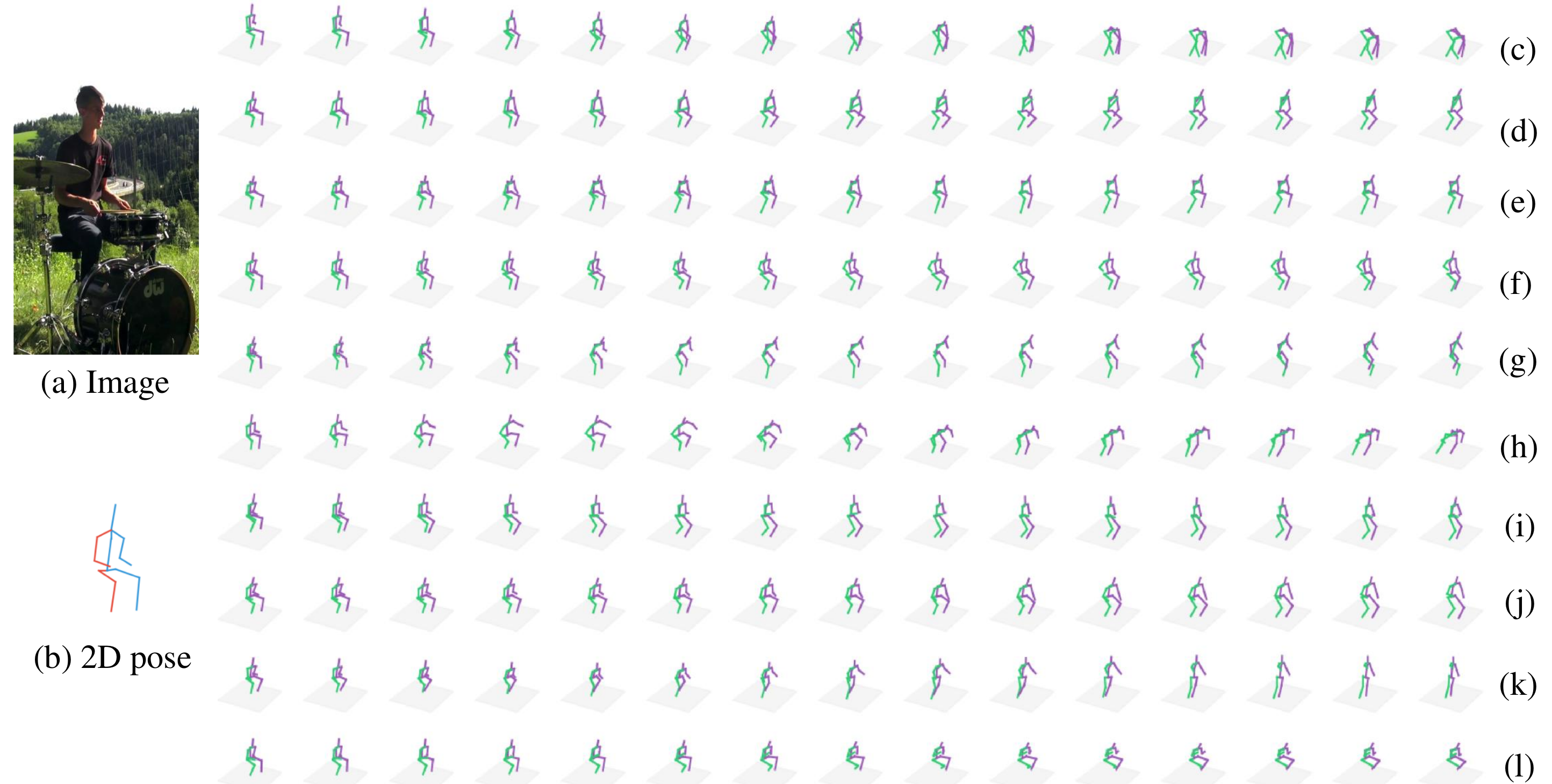}
\end{center}
\caption{ Example of diverse future motion prediction results ((c) $\sim$ (l)) on the input image ``Man sitting'' (a) in MPII.}
\label{fig:res_MPII_2}
\end{figure}

\begin{table}[t]
\centering
\caption
{Evaluation on MPJPE and APD under different types of settings. 2nd $\sim$ 4th columns display the influence of number of Gaussian components $M$, and the final columns shows the result of training without $L_{d}$ in Eq. \ref{eq:eq15}.  }
\label{tab:tab3}
\scalebox{1.0}{
\begin{tabular}{{lccc:c}}
\hline
      & $M$=7    & $M$=10          & $M$=12           & $M$=10 (w/o $L_d$) \\ \hline
MPJPE $\downarrow$ & 106.06 & 102.86        & \textbf{101.15} & 103.31           \\
APD $\uparrow$   & 2.70   & 2.85  & \textbf{2.92}           & 2.83            \\ \hline
\end{tabular}}
\end{table}

\subsection{Ablation Studies}
\label{sec:Ablation Studies}
To provide a deeper understanding of our method, we conduct ablation studies to analyze the prediction performance with respect to the following two aspects:

\noindent	\textbf{Number of Gaussian components $M$. }
We evaluate the prediction performance under different $M$
to gain more insights into its impact in terms of prediction accuracy and diversity. We adopt the Average Pairwise Distance (APD) metric for diversity measurement. APD averages the L2 distance between all the pairs of generated motion hypotheses, and a higher APD indicates a richer diversity within motions. The results on changes of MPJPE (accuracy) / APD (diversity) with different $M$ are reported in the 2nd $\sim$ 4th columns in Tab. \ref{tab:tab3}. It can be observed that a larger $M$ leads our model to better characterize both accurate and diverse predictions, while sacrificing more computational consumption. Also, a small $M$ contributes less to both accuracy and diversity. We empirically set $M$ to 10 in all experiments since there is no significant increase in accuracy in the case of a larger $M$ (i.e., $M=12$).

\vskip\baselineskip

\noindent	\textbf{Impact of $L_d, L_v$ and $L_a$.}
We introduce the energy-based formulation to maintain motion coherence ($L_v$) and boost prediction accuracy ($L_a$). Tab. \ref{tab:tab2} shows the MPJPE by including different losses. It can be observed from the 4th row in Tab. \ref{tab:tab2} that the prediction performance generally improves by adding $L_v$, including the two challenging cases of actions in ``Sitting'' and ``SittingDown''. In particular, for ``Sitting'' and ``SittingDown'', since most of the actions in Human3.6M are standing-based poses, $L_v$ is empowered to regularize inter-pose velocity such that motion coherence in large global movements (e.g., from standing to sitting) can also be retained. Besides, it is worth mentioning that further adding $L_a$ (last row in Tab. \ref{tab:tab2}) contributes to the best performance of our method. 

The adoption of $L_d$ is to regularize the mixture coefficients to alleviate the degeneration. We can see in Tab. \ref{tab:tab2} (3rd row) that training with $L_d$ improves the MPJPE specifically on the actions ``Discussion'' and ``Purchase''. We further investigate the APD and MPJPE without $L_d$ (i.e., only using $L_{MD},L_{v}$ and $L_a$ in Eq. \ref{eq:eq15}) and display the result in the last column in Tab. \ref{tab:tab3}. It can be confirmed that using $L_d$ also slightly refines both MPJPE and APD. The above analysis validates that $L_v, L_a$ and $L_d$ respectively possess the power to positively promote the prediction performance.

\section{Conclusion}
In this paper, we proposed a novel human motion prediction strategy by resorting to MDN modeling. Our model takes an image as input and handles the inner ambiguities in the weakly informed condition by generating diverse plausible future motion hypotheses. We also formulated an energy-based prior loss to boost the prediction performance. Extensive experiments on two standard datasets qualitatively and quantitatively demonstrated the generalization and capability of our approach in characterizing both accurate and diverse predictions.

Our method involves a major limitation in its dependence on the 2D motion extraction module. If the stacked hourglass network cannot guarantee a reasonable 2D pose, our model will fail to predict the accurate motion. Hence, we next would like to explore the possibilities of either adopting a more powerful 2D estimator or developing an end-to-end manner of training policy. In addition, our model is not able to recognize the most probable motion among the generated diverse predictions. In the future, we would also focus on resolving this issue by further learning a classification network to identify the best mode that \re{corresponds the most} to the ground-truth.


{\small
\bibliographystyle{cvm}
\bibliography{reference}
}

\end{document}